\documentclass[letterpaper, 10 pt, conference]{ieeeconf}  % Comment this line out
\IEEEoverridecommandlockouts                              % This command is only
\usepackage{graphicx}
\usepackage{url}
\usepackage{float}
\usepackage{subcaption}
\usepackage{amsmath}
\usepackage[utf8]{inputenc}
\usepackage{hyperref}
\usepackage{cite}

\title{\LARGE \bf
 ARED: Argentina Real Estate Dataset
}

\author{
    Iván Belenky \\% <-this % stops a space
    \href{mailto:ivan.belenky@ib.edu.ar}{\texttt{ivan.belenky@ib.edu.ar}}
}

\begin{document}

\maketitle
\thispagestyle{empty}
\pagestyle{empty}

%%%%%%%%%%%%%%%%%%%%%%%%%%%%%%%%%%%%%%%%%%%%%%%%%%%%%%%%%%%%%%%%%%%%%%%%%%%%%%%%
\begin{abstract}

The Argentinian real estate market presents a unique case study characterized by its unstable and rapidly shifting macroeconomic circumstances over the past decades. Despite the existence of a few datasets for price prediction, there is a lack of mixed modality datasets specifically focused on Argentina. In this paper, the first edition of ARED is introduced. A comprehensive real estate price prediction dataset series, designed for the Argentinian market. This edition contains information solely for Jan-Feb 2024. It was found that despite the short time range captured by this zeroth edition (44 days), time dependent phenomena has been occurring mostly on a market level (market as a whole). Nevertheless future editions of this dataset, will most likely contain historical data. Each listing in ARED comprises descriptive features, and variable-length sets of images. This edition is available at \url{https://github.com/ivanbelenky/ARED}. %Although individual habitable space prices exhibit time dependence, the real estate market in Argentina largely moves as a cohesive whole. Most models successfully capture the relative importance of features, thereby transforming the price prediction problem from a time-dependent snapshot model into a more straightforward scaling approach.

\end{abstract}

%%%%%%%%%%%%%%%%%%%%%%%%%%%%%%%%%%%%%%%%%%%%%%%%%%%%%%%%%%%%%%%%%%%%%%%%%%%%%%%%
\section{INTRODUCTION}
The Argentine economy has been marked by a series of economic crises throughout its history. The crisis of the 1970s, known as the "Rodrigazo," was named after the former Minister of Economy, who implemented shock measures such as the devaluation of the Argentine Peso and the increase in fuel prices. It has been argued \cite{gaggero2013origen} that this period marked the beginning of the dominance of the US dollar as a refuge of value in the Real Estate market, positioning itself as the primary currency for pricing and transactions. Since then, the housing market, particularly in terms of buying operations, has predominantly valued properties in USD. Access to credit for housing investments is limited, and most real estate transactions are conducted with cash in hand.\\

Furthermore, recent monetary policies, implemented from 2013 to the present (February 2024), have restricted access to foreign currencies. This has led to an outflow of pesos into the construction market, as investors seek to "dollarize" their assets through construction investments.\\

Current circumstances do not deviate from the traditional chaos-like economic scenarios for Argentina. The recently elected, self-proclaimed far-right government is presently implementing shock measures. Once again devaluing the Argentine peso, along with various other unpopular and orthodox economic policies. Their objective is to rectify, once and for all, as they say, the root causes of Argentina's economic disorder. This approach is being pursued with the view of rewriting Argentina's complex history in one fell swoop, underestimating the necessity for political consensus and risking the stability of the measures. Concerns that have been even echoed by the International Monetary Fund (IMF) during their visits to the country.\\

\begin{figure*}[t]
    \centering
    \includegraphics[width=\textwidth]{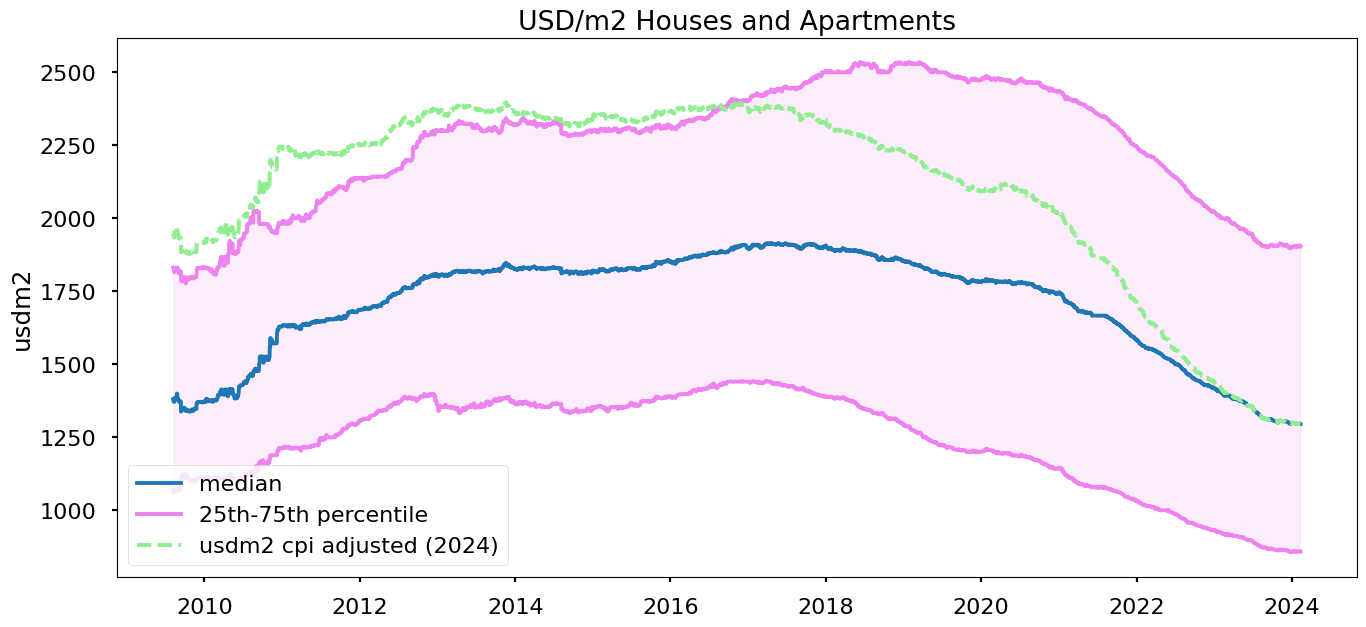}
    \caption{Price per square meter for houses and apartments across Argentina's Territory, 25th-75th quantile ranges and CPI adjustment median values.}
    \label{fig1}
\end{figure*}

\begin{figure*}[t]
    \centering
    \includegraphics[width=\textwidth]{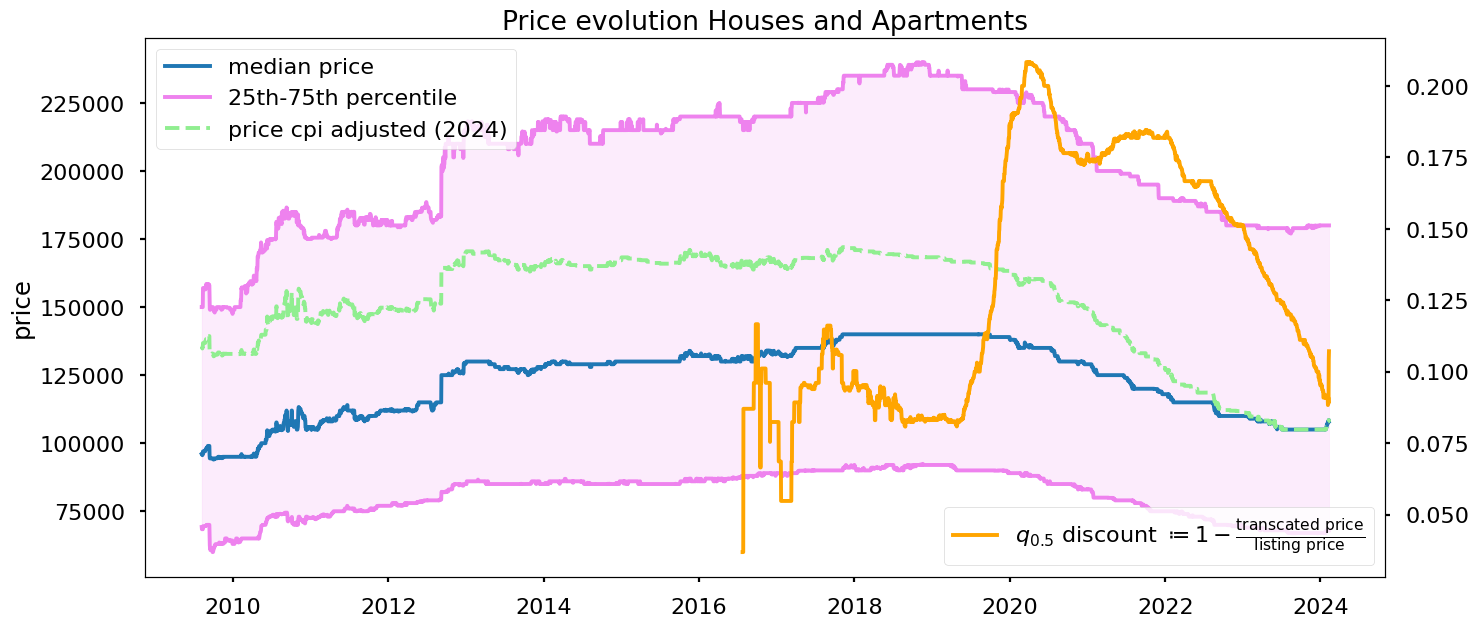}
    \caption{Price for houses and apartments across Argentina's Territory, 25th-75th quantile ranges and CPI adjustment median values. Discount rates are overlapped on the timeseries.}
    \label{fig2}
\end{figure*}

The real value of housing is undergoing a deep pricing valley, with no good outlook ahead for sellers in the near future, and great opportunity for investors and buyers. After certain stability for the past 15 years, the real estate market is experienced a deep decline across its territory, where the market moves pretty much as a whole (Fig. \ref{fig1} \& \ref{fig2}).\\

Historical and current events configure an interesting case study for the real estate market analysis, and price prediction problem. Despite the existence of some datasets focused on Latin America \cite{latamDS} there is no available time snapshot dataset consisting of multi modal information.\\

\section{The ARED dataset}

Motivated by the lack of multimodal data the ARED dataset (0th version) was constructed via 
\begin{itemize}
    \item automated listing scraping from major real estate companies
    \item automated image scraping and homogeneization
    \item post processing and pruning (data entry errors from the real estate agents)
\end{itemize}. 

\begin{figure}[H]
\centering
    \includegraphics[width=0.4\textwidth]{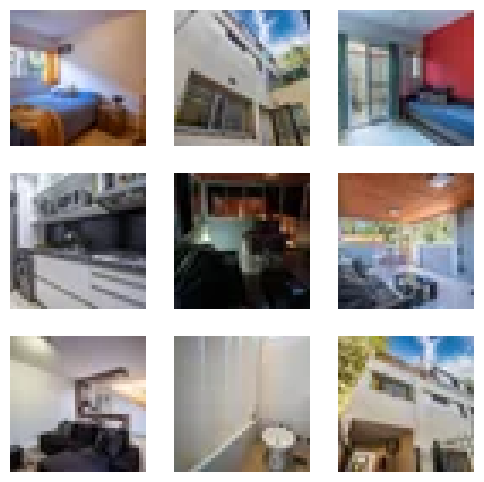}    
\end{figure}

The onset date for its creation is mid-january 2024 (11$^{\text{th}}$). The main goal of this is to expand it on a quarterly basis in order to build a comprehensive and tracking resource for the Argentinian real estate market. For each listing $l_{i}$ available in the dataset the following is included.
\begin{itemize}
\item $I_{i} := \{ \text{IMG}_{ij} \ \text{for} \ j \in [1, k_i]\}$ where $\text{IMG}_{ij}$ is an RGB channeled $40 \times 40$ image and $k_i \in [1,404]$.
\item listings features
    \begin{itemize}
        \item \textbf{id}: MD5 hash
        \item \textbf{property type}: $\in$ [Houses, Apartment, Land, Utilitary, Commercial Use, ...]. Currently, 26 different types exists. Though, most property types correspond to some form of apartment or houses. This habitable group contains approximately $~88\%$ of the dataset listings (table \ref{table1})
        \item \textbf{condition} $\in$ [\textbf{brand new}, \textbf{to refactor}, \textbf{in construction}, \textbf{excellent}, \textbf{good}, \textbf{very good}, \textbf{regular}]. This categories are traits selected by the real estate market agents so it is a vairable to be taken with a grain  of salt. There is no detrimental category such as BAD, so the subjectivity for this field should be taken into account when utilizing the dataset.
        
        \item \textbf{year built}
        \item \textbf{price} \& \textbf{price currency}.
        \item $[\textbf{latitude, longitude}]$
        \item \textbf{$m^{2}$ info $\in$ [$m^{2}$ total, $m^{2}$ covered, $m^{2}$ land]}
        \item \textbf{$\#$rooms, $\#$bedrooms, $\#$bathrooms, $\#$toilets}
        \item \textbf{l2}: administrative level 2 (province)
        \item \textbf{timestamps}: creation datetime, last seen datetime
        \item \textbf{description}: it is a free text field used mainly to describe the property and its features in natural language.
    \end{itemize} 
\end{itemize}

\begin{table}[h]
\fontsize{14}{14}{
\centering
\begin{tabular}{ll} 
\hline
\multicolumn{1}{l}{Property Type} & \multicolumn{1}{l}{\%}  \\
\hline
\hline
Apartments  Houses                  & 87.8\%                 \\ 
\hline
Land                                & 1.0\%                  \\ 
\hline
Commercial Use                      & 2.4\%                  \\ 
\hline
Others                              & 8.8\%                 
\end{tabular}
\caption{}
\label{table1}
}
\end{table}

\section{SNAPSHOT}
Despite the proved benefits from temporal data for price prediction \cite{lee2023strap, DBLP:journals/corr/PoursaeedMB17, wu2021autoformer, 10184898, xu2020spatial, zhou2021informer}  it must be stated that although individual habitable space prices exhibit time dependence, the real estate market in Argentina has been moving for the past 5 years as a cohesive whole. With this into consideration, ARED 0th edition, a 44 days wide snapshot, should still be useful to find determinants of value for listings.\\

\begin{figure}[ht]
    \includegraphics[width=0.5\textwidth]{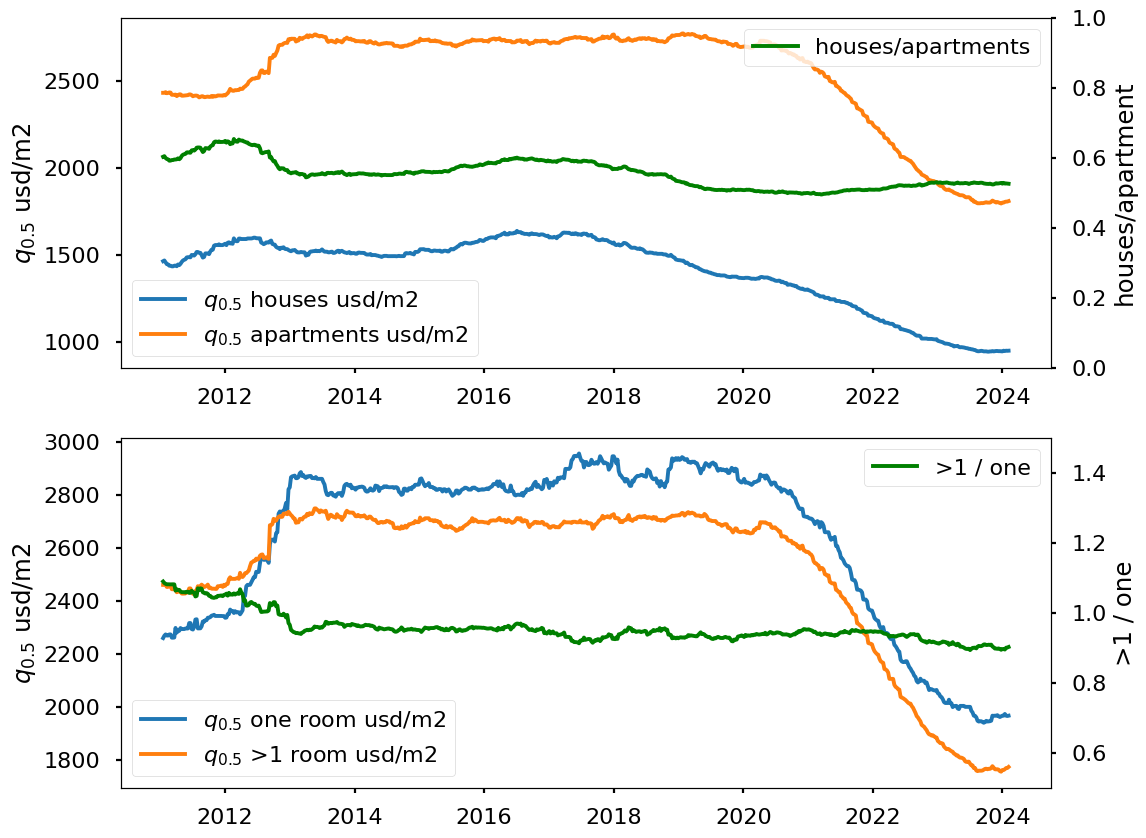}    
    \caption{Relative statistics between property type groups: houses and apartments \& 1 room apartments and $>$1 room apartments.}
    \label{fig4}
\end{figure}

The \textbf{maaw} (market as a whole) behavior exists for instance when houses and apartments median prices per square meter are compared (Fig \ref{fig4}).

\begin{figure}[ht]
    \includegraphics[width=0.5\textwidth]{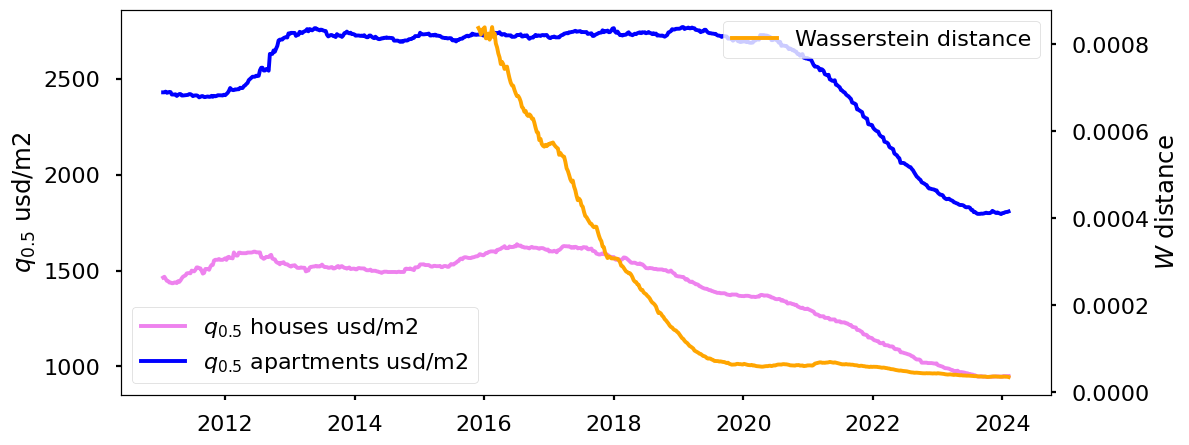}    
    \caption{Wasserstein distance evolution between house and apartments prices distributions. Both houses and apartments data were constructed taking into account all of their respective subcategories such as House Duplex, Apartment Loft, etc.}
    \label{fig5}
\end{figure}

Furthermore, not only the median statistics shows the \textbf{maaw} trend. If the Wasserstein distance between Houses and Apartment prices distribution is calculated across time, it can be seen (Fig \ref{fig5}) that it has stayed constant from the onset of the downard trends in listings prices. This would indicate that when the market is dropping, it drops cohesively. Also, given the reduction in the nominal value of the Wasserstein distance itself it could be argued that once again, all habitable spaces are more equal to each other in terms of relative valuation, reinforcing the idea that prices can only drop as a cohesive whole.\\

Cohesive behavior, constant relative prices and relative prices distribution should be a guarantee towards the usefulness of the first edition of ARED.\\

\section{FUTURE OF ARED}

\begin{figure}[ht]
    \includegraphics[width=0.5\textwidth]{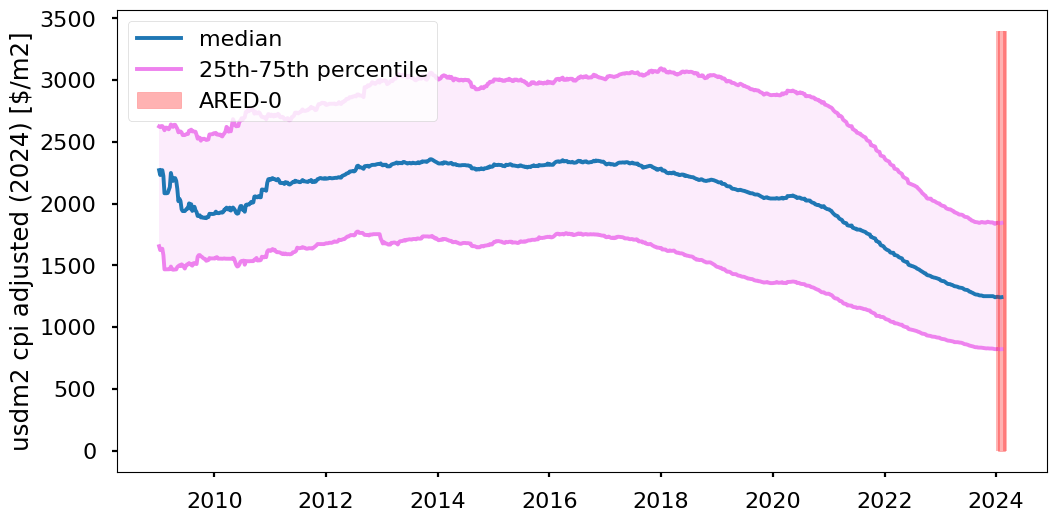}    
    \caption{Visual ARED0 range relative to the past historical data.}
    \label{fig6}
\end{figure}

ARED commits to regular updates, with new editions scheduled to be released on a quarterly basis, next one expected on mid May 2024. In addition to updating the dataset with fresh data, there are plans to enrich ARED by including historical data preceding the 11th of January 2024. This historical data will provide valuable context and insights into long-term trends and patterns within the Argentinian real estate market (Fig \ref{fig6}).\\

\addtolength{\textheight}{-12cm}   % This command serves to balance the column lengths

%%%%%%%%%%%%%%%%%%%%%%%%%%%%%%%%%%%%%%%%%%%%%%%%%%%%%%%%%%%%%%%%%%%%%%%%%%%%%%%%

%%%%%%%%%%%%%%%%%%%%%%%%%%%%%%%%%%%%%%%%%%%%%%%%%%%%%%%%%%%%%%%%%%%%%%%%%%%%%%%%

%%%%%%%%%%%%%%%%%%%%%%%%%%%%%%%%%%%%%%%%%%%%%%%%%%%%%%%%%%%%%%%%%%%%%%%%%%%%%%%%

%\begin{thebibliography}{99}

%\bibitem{gaggero2013origen}
%  \textsc{Gaggero, A.}, \& \textsc{Nemi\~{n}a, P.} (2013). El origen de la dolarizaci{\'o}n inmobiliaria en la Argentina. \textit{Sociales en debate}, (5).

%\bibitem{latamDS}
%  \textsc{Jacobsen, Rasmus}, A. (2020). Property Listings for 5 South American Countries. \textit{Kaggle}. Retrieved from \url{https://www.kaggle.com/datasets/rmjacobsen/property-listings-for-5-south-american-countries/data}
%\bibitem

%\bibitem{lee2023strap}
 %   Hojoon Lee, Hawon Jeong, Byungkun Lee, Kyungyup Daniel Lee, and Jaegul Choo.
  %  \newblock ST-RAP: A Spatio-Temporal Framework for Real Estate Appraisal.
%    \newblock In \textit{Proceedings of the ACM International Conference on Information and Knowledge Management (CIKM)}, 2023.
%\bibitem
\bibliographystyle{plain}
\bibliography{bib}

\begin{thebibliography}{1}

\bibitem{gaggero2013origen}
Alejandro Gaggero and Pablo Nemi{\~n}a.
\newblock El origen de la dolarizaci{\'o}n inmobiliaria en la argentina.
\newblock {\em Sociales en debate}, (5), 2013.

\bibitem{latamDS}
Rasmus~A. Jacobsen.
\newblock Property listings for 5 south american countries.
\newblock \url{https://www.kaggle.com/datasets/rmjacobsen/property-listings-for-5-south-american-countries/data}, 2020.

\bibitem{lee2023strap}
Hojoon Lee, Hawon Jeong, Byungkun Lee, Kyungyup~Daniel Lee, and Jaegul Choo.
\newblock St-rap: A spatio-temporal framework for real estate appraisal.
\newblock In {\em Proceedings of the ACM International Conference on Information and Knowledge Management (CIKM)}, 2023.

\bibitem{DBLP:journals/corr/PoursaeedMB17}
Omid Poursaeed, Tomas Matera, and Serge~J. Belongie.
\newblock Vision-based real estate price estimation.
\newblock {\em CoRR}, abs/1707.05489, 2017.

\bibitem{wu2021autoformer}
Haixu Wu, Jiehui Xu, Jianmin Wang, and Mingsheng Long.
\newblock Autoformer: Decomposition transformers with auto-correlation for long-term series forecasting.
\newblock {\em Advances in Neural Information Processing Systems}, 34:22419--22430, 2021.

\bibitem{10184898}
Congxi Xiao, Jingbo Zhou, Jizhou Huang, Hengshu Zhu, Tong Xu, Dejing Dou, and Hui Xiong.
\newblock A contextual master-slave framework on urban region graph for urban village detection.
\newblock In {\em 2023 IEEE 39th International Conference on Data Engineering (ICDE)}, pages 736--748, 2023.

\bibitem{xu2020spatial}
Mingxing Xu, Wenrui Dai, Chunmiao Liu, Xing Gao, Weiyao Lin, Guo-Jun Qi, and Hongkai Xiong.
\newblock Spatial-temporal transformer networks for traffic flow forecasting.
\newblock {\em arXiv preprint arXiv:2001.02908}, 2020.

\bibitem{zhou2021informer}
Haoyi Zhou, Shanghang Zhang, Jieqi Peng, Shuai Zhang, Jianxin Li, Hui Xiong, and Wancai Zhang.
\newblock Informer: Beyond efficient transformer for long sequence time-series forecasting.
\newblock In {\em Proceedings of the AAAI conference on artificial intelligence}, volume~35, pages 11106--11115, 2021.

\end{thebibliography}

%\end{thebibliography}

\end{document}